# LLM-based Extraction of Contradictions from Patents

Stefan Trapp[1(✉)], Joachim Warschat[1]

[1] University of Hagen, Universitätsstraße 47, 58097 Hagen, Germany
stefan.trapp@fernuni-hagen.de

**Abstract.** Already since the 1950s TRIZ shows that patents and the technical contradictions they solve are an important source of inspiration for the development of innovative products. However, TRIZ is a heuristic based on a historic patent analysis and does not make use of the ever-increasing number of latest technological solutions in current patents. Because of the huge number of patents, their length, and, last but not least, their complexity there is a need for modern patent retrieval and patent analysis to go beyond keyword-oriented methods. Recent advances in patent retrieval and analysis mainly focus on dense vectors based on neural AI Transformer language models like Google BERT. They are, for example, used for dense retrieval, question answering or summarization and key concept extraction. A research focus within the methods for patent summarization and key concept extraction are generic inventive concepts respectively TRIZ concepts like problems, solutions, advantage of invention, parameters, and contradictions. Succeeding rule-based approaches, finetuned BERT-like language models for sentence-wise classification represent the state-of-the-art of inventive concept extraction. While they work comparatively well for basic concepts like problems or solutions, contradictions − as a more complex abstraction − remain a challenge for these models. Even PaTRIZ, the latest and complicated multi-stage approach to extract contradictions, delivers only mixed results. This paper goes one step further, as it presents a method to extract TRIZ contradictions from patent texts based on Prompt Engineering using a generative Large Language Model (LLM), namely OpenAI's GPT-4. The existing annotated patent dataset "PaGAN" is used to demonstrate the LLM-capabilities for extracting TRIZ contradictions from the section "State-of-the-Art" of USPTO patents. Contradiction detection, sentence extraction, contradiction summarization, parameter extraction and assignment to the 39 abstract TRIZ engineering parameters are all performed in a single prompt using the LangChain framework. Our results show that "off-the-shelf" GPT-4 is a serious alternative to PaTRIZ. Comparing the text similarity of the GPT-4 extractions with the annotated sentences from PaGAN we reach a high F1-value of 0.93 using the BERTScore metric.

## 1 Introduction

Innovations are a key success factor for companies to compete in the age of globalization. Emerging technologies are an important driver for innovative products as they provide new or improved solutions for customer needs. But as the technical knowledge published worldwide grows rapidly and technologies become increasingly complex, it gets harder for companies to identify attractive technological opportunities especially outside their own domain. This implies the risk of missing important technological developments and of falling behind in competition. [1]

Because 85 to 90 percent of the worldwide technical knowledge can be found in patents, they are an extensive and cross-domain knowledge resource for inventive technological solutions [2]. This potential of patents for improved innovation results was already discovered by Altshuller in the 1950s, when he developed TRIZ, the theory of inventive problem solving [3]. Back then patent search engines did not yet exist, but Altshuller's manual patent analysis revealed two important insights. First, the resolution of contradictions between system parameters often creates highly innovative solutions. Second, many high-grade innovations are based on the cross-domain transfer of already existing solutions. Until today TRIZ remains a classic approach promising technical solutions of higher innovativeness than other general-purpose creativity techniques. TRIZ is based on a historic patent analysis, but it does not make use of the ever-increasing number of latest technological solutions in current patents. [4]

Nowadays, patent search engines support the retrieval of patents (possibly) containing technological solutions of interest. However, especially when it comes to cross-domain search, the widespread keyword-based lexical patent search suffers from the vocabulary mismatch problem: due to different terms, synonyms or a completely different jargon between domains, the word-overlap may only be very little, resulting in unsatisfactory search results.

A possible solution to simultaneously make use of the TRIZ findings and to overcome the problems of conventional lexical patent search is the extraction of inventive (key) concepts from patents and their subsequent use for patent retrieval. Generic inventive concepts present in all patents are "problems", "advantages of the invention" and "solutions" [5]. TRIZ adds the entities "parameter" and "contradiction" [4]. Among these inventive concepts contradictions are the most complex concept, and, according to TRIZ, also the most important one to find highly innovative solutions, if a solution is presented to overcome a contradiction.

Existing solutions for inventive concept extraction rely either on rule-based natural language processing (NLP) or, lately, on Artificial Intelligence (AI) in the form of pre-trained neural language models like Bidirectional Encoder Representations from Transformers (BERT) [6] developed by Google. After additional finetuning with annotated data, BERT and its derivatives are typically used for sentence-wise classification to extract inventive concepts. However, the automatic detection of contradictions in patent text remains a challenge.

Generative Large Language Models (LLMs), like OpenAI's omnipresent GPT (Generative Pre-trained Transformer) [7], are the latest trend in pre-trained neural language models. With their improved capabilities to understand and generate text LLMs often deliver satisfying results without effortful finetuning and simply via instructions and input in plain natural language (prompts and Prompt Engineering). Thus, they might also be an alternative to smaller models like BERT for inventive concept extraction.

The aim of this paper is to analyze the extraction of technical contradictions from patents in English language using "out-of-the-box" LLMs and Prompt Engineering. The LLM results are compared with the state-of-the-art, namely sentence classification with finetuned BERT derivatives and complicated multi-stage approaches based on the same.

## 2 Background and Related Work

### 2.1 Patents and their Inventive Concepts

Over 20 million patents exist worldwide. Their average text length is around 10,000 words. In addition to their length, their complicated writing style, with the intention to hide the intellectual property as much as possible, causes the vocabulary mismatch problem for keyword-oriented patent search. Contextual text embeddings based on neural networks avoid the dependency on words and word-overlap as they code the text's meaning. Therefore, they are a potential workaround for the vocabulary mismatch problem, which hinders particularly cross-domain search. The latest generation of such neural networks are transformer architectures like Google BERT. [8]

According to the guidelines of the World Intellectual Property Organization (WIPO), patents should cover the problem to be solved by the invention, the solution to overcome the problem and the advantageous effects of the invention [5]. Thus, these three generic inventive concepts can be expected to be present in all patents. The problems or technical contradictions addressed in a patent are often presented in its "Background" section describing the state-of-the-art [9].

### 2.2 TRIZ and its Inventive Concepts

The probably most famous and widespread tool of the TRIZ heuristic is the contradiction matrix for solving technical conflicts between parameters. The TRIZ-user must abstract the targeted and the deteriorated parameters of the problematic contradiction to one of the 39 TRIZ engineering parameters to select a cell of the contradiction matrix, thus obtaining a subset of the 40 generic TRIZ innovative principles as guideline for further solution search. These innovative principles proved to be successful for this class of problems in Altshuller's historic patent analysis and can be used to look for solutions adaptable to one's own problem. [10]

Parameters and contradictions can be seen as further generic inventive concepts in addition to problems, solutions, and advantageous effects. Contradictions between system parameters are a specific type of problems. According to TRIZ, overcoming contradictions promises excellent chances for highly innovative solutions.

Later developments based on TRIZ principles, like the Inventive Design Method (IDM) formalize TRIZ concepts through ontologies, enabling automated text analysis. IDM defines TRIZ contradictions as a set of two evaluation parameters (corresponding to the engineering parameters in the rows and columns of the TRIZ contradiction matrix) and possibly a third parameter, the action parameter, corresponding to the "physical contradictions" of classic TRIZ. [4]

### 2.3 Extraction of Inventive Concepts from Patents

Many approaches targeting the extraction of inventive concepts from patents rely on rule-based NLP (syntactics of sentences and signaling markers) to retrieve entities and to link them to each other [11, 12]. However, Heller and Warschat show that BERT outperforms static word embeddings such as Word2vec and GloVe in the sentence-wise classification of problems and solutions [13].

Similar to Heller and Warschat, Giordano et al. fine-tune a BERT-for-Patents model [14] for classifying patent sentences into the four classes "problem", "solution", "advantage (of the invention)" or "other sentences" [5].

Guarino et al. use two BERT-based binary sentence classifiers in their "SummaTRIZ" approach to extract sentences from patents that define contradictions [9]. "SummaTRIZ" was later supplemented with further NLP steps to form the "PaTRIZ" approach [15, 16, 17].

"PaTRIZ" is probably the most advanced contradiction extraction method identified in the state-of-the-art [15, 16, 17]. It is a complicated four-stage approach. Stage (1) is a “yes/no” classifier for the presence of contradictions in the respective patent. Stage (2) uses (like "SummaTRIZ") two binary BERT classifiers to extract those sentences that describe a contradiction. The first sentence classifier identifies sentences with the parameter to be improved ("first part of the contradiction"), the second identifies those with the deteriorating parameter ("second part of the contradiction"). The separate classification aims to find contradictions in which both components are contained in the same sentence [17]. Stage (3) is a Conditional Random Field (CRF) for extracting the parameters from the "contradiction sentences". In stage (4) an MPNet transformer classifies the extracted (concrete) parameters as one of the 39 abstract engineering parameters of the (classic) TRIZ contradiction matrix. This classification is based on the semantic text similarity with the textual description of the TRIZ parameters.

Despite its complexity and high finetuning effort the best PaTRIZ overall model only achieves a precision of 0.33, i.e. out of three contradictions extracted, only one matches to an expert-annotated contradiction. The classifier stage (2) alone reaches F1-values of 0.37 for the "first part" resp. F1 = 0.65 for the "second part”. [17]

### 2.4 Finetuning versus Prompt Engineering and In-Context-Learning

Finetuning is an approach for improving the results of pre-trained language models on a downstream task using an additional task-specific dataset. In contrast to other techniques like tuning the prompts for querying the model (Prompt Engineering), finetuning adapts the model weights (parameters) of the pre-trained language model. Supervised learning of neural models requires comprehensive training data, but this data is often difficult to obtain or difficult to annotate. Specifically in the patent domain there is a lack of such data. [18, 19]

Larger language models usually achieve significant performance improvements compared to smaller models, which can be described by so-called scaling laws [20, 21]. BERT with its 110 million (BERT-base) or 340 million (BERT-large) parameters is nowadays only considered a "medium-sized language model" [22] and is often used for natural language understanding tasks like text classification and keyword extraction. Since BERT, as a masked language model, was not designed to be generative (encoder architecture), it is considered less suitable for language generation tasks compared to generative models such as GPT. However, with its 175 billion parameters, GPT-3 is almost 1,600 times larger than BERT-base. With the successor GPT-4, the size has increased sixfold again, to around one trillion parameters [23]. Currently only GPT and comparable models with their greatly improved abilities in terms of language understanding, language generation, generalization, problem solving and the derivation of logical conclusions (logical reasoning) are considered as "Large Language Models" (LLMs). [24, 25]

For medium-sized language models such as BERT and its derivatives, finetuning is usually essential to achieve satisfactory results. On the other side, LLMs can be finetuned, too (with appropriately powerful hardware), but in many cases they deliver satisfying results without effortful finetuning and simply via instructions and input in plain natural language (prompts and Prompt Engineering). Prompts and Prompt Templates allow Few-shot learning, i.e. learning based on a selection of solved examples, without adjusting model parameters. This approach is also known as in-context learning (ICL). In many cases the thorough design of so-called zero-shot prompts can achieve comparable performance even without inserted examples [26, 27].

With their avoidance of finetuning and the intuitive codeless interaction via prompts LLMs might be an interesting alternative to smaller language models like BERT for inventive concept extraction and specifically for technical contradictions.

# 3 Proposed Method

This chapter describes the new, LLM-based method to extract technical contradictions from patent texts.

## 3.1 Dataset and Data Preparation

In this paper we reuse a patent corpus (called "PaGAN dataset" hereafter) provided by a research team of the Institut national des sciences appliquées de Strasbourg (INSA). It contains 3,200 English-language patents from the database of the United States Patent and Trademark Office (USPTO). Half of these patents (1,600) contain expert-annotated TRIZ contradictions, usually consisting of at least two sentences ("first part of the contradiction" and "second part of the contradiction"). The other half of the corpus is supposed to contain no contradictions. [28]

The PaGAN dataset is delivered in multiple text files per patent. Each patent section is stored in a separate file, plus (if applicable) a text file for the expert-annotated sentences describing the contradiction. This format makes the data access cumbersome and slow for batch-use in combination with language models. Thus, a data preparation step was performed. The PaGAN text files were "pickled" into a single Python object serialization file using the Pandas library (https://pandas.pydata.org/).

## 3.2 LLM-based Contradiction Extraction with LangChain and GPT-4

LangChain is a Python framework to create LLM-powered applications [29]. It can be used in cloud-based development environments like Google Colab [30]. Colab allows the easy execution of Python code as Jupyter Notebooks on GPUs with high computing capacities, thus the environment is particularly suitable for training and executing neural language models.

LangChain is prepared to use various open-source or commercial LLMs via their Application Programming Interface (API). One of these LLMs is OpenAI's GPT. In its version GPT-4 it is OpenAI's most advanced system and benchmarks consider it as the probably most powerful LLM currently available [7, 31]. We use GPT-4 Turbo release *gpt-4-1106-preview*, presented by OpenAI on 6 November 2023, to process the following prompt for contradiction extraction. LangChain's API-agnostic role "System" sets the objectives the AI should follow. We use it to explain the term "TRIZ contradiction" and give definitions for the 39 engineering parameters of the classic TRIZ contradiction matrix. The LangChain role "Human" sends messages from the perspective of the human user. We here define the tasks to be performed by the AI, request an answer in JSON (JavaScript Object Notation), specify its exact format, and, finally, present the text to be analyzed. For the sake of brevity the TRIZ engineering parameters 2 to 38 are omitted in the prompt below.

```
System:
As a worldclass patent and technology expert you are extracting TRIZ contradictions
from text passages of patents.
A TRIZ contradiction is indicated by at least 2 conflicting parameters, attributes
or properties that cannot be optimized simultaneously.
These parameters are called 'evaluation parameters'. If one of these parameters is
improved, the other one deteriorates and vice versa.
Possibly a third parameter is part of the contradiction, which influences both
'evaluation parameters' in opposite directions. This optional third parameter is
called 'action parameter'. The 'action parameter' must be maximized to improve one
of the 'evaluation parameters' while it must be minimized to improve the other.
Each 'evaluation parameter' can be classified as one of the 39 abstract TRIZ-
parameters of the TRIZ contradiction matrix given below.
'1 -  Weight of Moving Object': The mass of the object, in a gravitational field.
      The force that the body exerts on its support or suspension.
// ...TRIZ-parameters 2-38
'39 - Productivity': The number of functions or operations performed by a system
      per unit time. The time for a unit function or operation. The output per unit
      time, or the cost per unit output.

Human:
Perform the following tasks on the given text:
1. Decide if the text mentions at least one TRIZ contradiction. Your answer must be
   either 'True' or 'False'. Answer only 'True', if the conflicting parameters are
   apparent. 'False' otherwise.
```

```
If the answer to task 1 is 'True', perform additionally the following tasks per
contradiction mentioned:
2. Quote without any modification the sentences containing the parameters which
   need to be improved respectively involved in the initial problem. Quote a
   maximum of 6 sentences. Order them by decreasing importance. Avoid duplicate
   sentences for this task and task 3 below.
3. Quote without any modification the sentences containing the draw backs of the
   prior art solutions. Quote a maximum of 6 sentences. Order them by decreasing
   importance. Avoid duplicate sentences for this task and task 2 above.
4. Summarize the TRIZ contradiction in one sentence WITHOUT any introductory phrase
   like 'The TRIZ contradiction is that ...'. Include values and units for the
   parameters, if given.
5. Analyze the parameters of the TRIZ contradiction. Return the 'evaluation
   parameters' and, if mentioned, the 'action parameter'.
6. Assign each of the 'evaluation parameters' to one of the 39 TRIZ-parameters of
   the TRIZ contradiction matrix. Quote solely the text in inverted commas from
   their description, nothing else.

Format your response in a structured JSON-like format, as follows:

{"Contradictions": [{
  "is_contradiction": "True/False",
  "target_sentences": "['Only if 'is_contradiction' is True: quote without any
  modification the sentences containing the parameters which need to be improved
  respectively involved in the initial problem. Quote a maximum of 6 sentences.']",
  "drawback_sentences": "['Only if 'is_contradiction' is True: quote without any
  modification the sentences containing the drawbacks of the prior art solutions.
  Quote a maximum of 6 sentences.']",
  "summary": "summary, only if 'is_contradiction' is True",
  "evaluation_parameter_1": "parameter, only if 'is_contradiction' is True",
  "evaluation_parameter_2": "parameter, only if 'is_contradiction' is True",
  "triz_parameter_1": "TRIZ parameter, only if 'is_contradiction' is True",
  "triz_parameter_2": "TRIZ parameter, only if 'is_contradiction' is True",
  "action_parameter": "parameter, only if 'is_contradiction' is True and the
  action_parameter exists",
},
// ... additional contradictions, if present
]}

Return the JSON code only.

The text to analyze is: ...
```

## 4 Evaluation

The results of the LLM-based contradiction extraction are evaluated in this chapter in a qualitative and a quantitative way. For the qualitative evaluation we present two patents as case study. The quantitative evaluation relies on the portion of correctly extracted sentences compared with the PaGAN annotations and on the semantic text similarity measured with the F1-value of the BERTScore metric.

### 4.1 Case Study

The Patents US06938300 "Wheel assembly for a stroller" and US07543344 "Cover for a heating blanket" are used as cases to evaluate the proposed LLM contradiction extraction. The section's "State-of-the-Art" content from US06938300 is shown below. Contradiction sentences according to PaGAN dataset are annotated in **bold**:

```
1. This invention relates to a stroller, and more particularly to a wheel assembly
for a stroller, which includes a single wheel. 2. Referring to 1 and 2, a conven-
tional stroller 1 is shown to include a stroller frame 10 with four legs 101, two
front wheel assemblies 11 mounted respectively to two of the legs 101, and two rear
wheel assemblies 12 mounted respectively to the other two of the legs 101. 2 is a
bottom view of one of the front wheel assemblies 11. Each of the front wheel assem-
blies 11 includes two front wheels 13, a wheel seat 14 disposed between the front
wheels 12, and a wheel axle 15 extending through the front wheels 13 and the wheel
```

```
seat 14. Each of the wheel seats 14 has a leg-connecting portion 141 sleeved rotat-
ably on a respective one of the legs 101, and an axle-connecting portion 142, with-
in which the wheel axle 15 is journalled. To enhance the comfort of the baby car-
ried on the stroller 1, a vibration-absorbing device not shown can be disposed
within the axle-connecting portion 142 of each of the wheel seats 14. When it is
desired to push the stroller 1 to advance along a straight path, a forward force A
is applied to the stroller frame 10. In each of the front wheel assemblies 11,
since the forward force A is located midway between two frictional forces B that
are generated between the ground and the front wheels 13 and since the direction of
the forward force A is parallel to those of the frictional forces B, the stroller 1
can advance along a straight path 16. When the stroller 1 moves over a lawn or une-
ven road surfaces, it is necessary for the stroller wheels to have a large diameter
so as to ensure the comfort of the baby. However, if each of the front wheel assem-
blies 11 has two large-diameter front wheels 13, the total volume and weight of the
stroller 1 will increase significantly so that it is difficult to push the stroller
1.
```

Feeding this patent section into the LLM *gpt-4-1106-preview* using LangChain and the prompt described in chapter 3 leads to the following AI-answer in JSON:

```
{
  "Contradictions": [
    {
      "is_contradiction": "True",
      "target_sentences": [
        "When the stroller 1 moves over a lawn or uneven road surfaces, it is nec-
        essary for the stroller wheels to have a large diameter so as to ensure the
        comfort of the baby."
      ],
      "drawback_sentences": [
        "However, if each of the front wheel assemblies 11 has two large-diameter
        front wheels 13, the total volume and weight of the stroller 1 will in-
        crease significantly so that it is difficult to push the stroller 1."
      ],
      "summary": "Large-diameter wheels are needed for baby comfort on uneven sur-
      faces, but increase the stroller's volume and weight, making it difficult to
      push.",
      "evaluation_parameter_1": "comfort of the baby",
      "evaluation_parameter_2": "total volume and weight of the stroller",
      "triz_parameter_1": "'Convenience of Use'",
      "triz_parameter_2": "'Weight of Moving Object'",
      "action_parameter": "diameter of the stroller wheels"
    }
  ]
}
```

In this case both sentences of the contradiction are extracted correctly. The GPT-4-generated summary covers all aspects of the contradiction. Evaluation parameters, action parameter and the assigned generic TRIZ engineering parameters are also correct. Altogether GPT-4's answer can be deemed as very good.
As a second case we use the patent US07543344 ("Cover for a heating blanket"). The content of the "State-of-the-Art" section is omitted here for brevity. It consists of a total of 3008 characters in 21 partially very long sentences. GPT-4's result is shown below.

```
{
  "Contradictions": [
    {
      "is_contradiction": "True",
      "target_sentences": [
        "FAW has become the standard of care for preventing and treating the hypo-
        thermia caused by anesthesia and surgery.",
        "There is a need for heating blanket covers that are easily manipulated for
        placing a heating blanket therein and do not compromise the comfort of a
        person over which the blanket is placed."
      ],
      "drawback_sentences": [
```

```
        "Although FAW is clinically effective, it suffers from several problems
        including: a relatively high price; air blowing in the operating room,
        which can be noisy and can potentially contaminate the surgical field; and
        the inflatable blanket is relatively bulky over the patient, at times even
        obscuring the view of the surgeon.",
        "Moreover, the low specific heat of air and the rapid loss of heat from air
        requires that the temperature of the air, as it leaves the hose, be danger-
        ously highin some products as high as 45 C.",
        "This creates significant dangers for the patient.",
        "Cleaning is not only time consuming during the rapid turnover of the oper-
        ating room after each case, but the labor for the cleaning is also expen-
        sive."
      ],
      "summary": "The need to prevent hypothermia in surgical patients with FAW
      conflicts with the drawbacks of high price, noise, potential contamination,
      bulkiness, and safety hazards.",
      "evaluation_parameter_1": "comfort of a person over which the blanket is
      placed",
      "evaluation_parameter_2": "safety hazards",
      "triz_parameter_1": "'33 - Convenience of Use'",
      "triz_parameter_2": "'30 - Harmful Factors Acting on Object'",
      "action_parameter": "temperature of the air"
    }
  ]
}
```

According to the PaGAN dataset 6 sentences are part of the contradiction. GPT-4 also identifies 6 sentences, but only 2 of them match with PaGAN. This result does not look very promising at a first glance. However, content-related (human) analysis shows that GPT-4's results are indeed plausible. The pros and cons of "forced-air warming" (FAW) are described in all 6 sentences. Furthermore, summary and all parameters including the assigned TRIZ engineering parameters seem correct.

The case study shows that the extraction (classification) of sentences for contradiction-identification may not be possible as precisely as desired, because of the complexity and fuzziness of patent language. GPT-4 as an example of state-of-the-art LLMs shows promising abilities not only to extract such sentences, but also to summarize technical contradictions, to identify their parameters and to assign them to the generic TRIZ engineering parameters.

### 4.2 Quantitative Evaluation

Out of the 1,600 patents annotated in the PaGAN dataset as patents **with** contradictions, an arbitrary selection of 100 patents was fed into the LLM *gpt-4-1106-preview* using *LangChain*. **Table 1** shows the results. In each of the 100 patents GPT-4 found at least one contradiction. In one of the patents two contradictions were identified.

The average number of annotated sentences in the PaGAN dataset is 3.81. GPT-4 identifies 2.69 or 71 % of these correctly (*Recall* = 0.71), 1.12 (29 %) are not found (false negative). On average 3.19 sentences are false positive, i.e. identified by GPT-4, but not annotated in PaGAN. Consequently, *Precision* is 0.46 and F1 is 0.56. The best PaTRIZ part models for sentence classification reach $F1_1 = 0.37$ ("first part" based on 2,276 positive sentences) resp. $F1_2 = 0.65$ ("second part" based on 3,709 negative sentences). The weighted average of $F1_1$ and $F1_2$ is 0.54. Based on our sample of 100 patents it can thus be concluded that a zero-shot *gpt-4-1106-preview* (F1 = 0.56) is at least on par with (if not slightly better than) the best performing fine-tuned sentence classifiers from PaTRIZ (F1 = 0.54).

As the PaGAN dataset provides the contradiction-involved sentences only "as is" and not a summary of the contradiction, these sentences are the basis for our quantitative similarity evaluation. We use the F1-value of the BERTScore metric for text similarity [32] and its Huggingface implementation (https://huggingface.co/spaces/evaluate-metric/bertscore). The metric's default model for the English language is "roberta-large". Comparing the "PaGAN sentences" with the concatenated GPT-4 sentences results in BERTScore-F1 = 0.93. The comparison with the GPT-4-generated summary delivers a slightly lower BERTScore-F1 = 0.88. Both values indicate a high similarity between the PaGAN annotations and the GPT-4 results.

**Table 1.** Results based on 100 patent texts **with** PaGAN contradictions

| **Metric** | |
|---|---|
| Patents with contradictions | 100/100 |
| Number of contradictions found | 101 |
| Mean number of sentences given in PaGAN dataset | 3.81 |
| Mean number of sentences extracted | 5.88 |
| Thereof: correctly extracted sentences (true positive) | 2.69 |
| Thereof: wrongly extracted sentences (false positive) | 3.19 |
| Mean number sentences not found (false negative) | 1.12 |
| Precision | 0.46 |
| Recall | 0.71 |
| F1 | 0.56 |
| BERTScore-F1 for concatenated sentences vs. PaGAN sentences | 0.93 |
| BERTScore-F1 for summary vs. PaGAN sentences | 0.88 |

Out of the 1,600 patents annotated in the PaGAN dataset as patents **without** contradictions, an arbitrary selection of 100 patents was fed into the LLM *gpt-4-1106-preview* using *LangChain*.

**Table 2.** Results based on 100 patent texts **without** PaGAN contradictions

| **Metric** | |
|---|---|
| Patents with contradictions | 86/100 |
| Number of contradictions found | 87 |
| Mean number of sentences given in PaGAN dataset | 0.00 |
| Mean number of sentences extracted | 4.84 |

**Table 2** shows that GPT-4 extracted contradictions from 86 of 100 patents, though the human PaGAN annotators tagged all 100 patents as "contradiction-free". For one of the 86 patents even two contradictions were found by GPT-4. With a mean of 4.84, the number of extracted sentences per patent is slightly lower than in the patents with PaGAN contradictions (5.88). One possible explanation for the discrepancy to the PaGAN annotations is that GPT-4 is not good at distinguishing between patents with and without contradictions. An alternative explanation is that the annotations in the PaGAN dataset are not correct, i.e. that "contradiction-free" patents in fact do contain contradictions. A decision between both possible explanations is not finally taken here, but from our perspective the following example of the patent "Golf ball utilizing silicon materials" (US07384349) indicates that at least some of the supposedly "contradiction-free" patents are not annotated correctly in the PaGAN dataset. We find the contradiction extracted by GPT-4, the summary, the parameters ("resilience or carrying distance", "playability characteristics", "multi-layer covers") and the generic TRIZ engineering parameters assigned ("15 - Durability of Moving Object", "33 - Convenience of Use") convincing.

```
{
  "Contradictions": [
    {
      "is_contradiction": "True",
      "target_sentences": [
        "The present invention is directed to new golf ball compositions, prefera-
        bly utilized in conjunction with multi-layer covers, which provide for en-
        hanced coefficient of restitution i.e, enhanced resilience or carrying dis-
        tance and/or durability properties when compared to the balls found in the
        prior art.",
        "Despite the great numbers of different materials and combinations of mate-
        rials utilized in prior art golf balls, there still remains a need for an
        improved golf ball exhibiting superior properties."
      ],
      "drawback_sentences": [
        "As such, the playability characteristics i.e., feel, click, spin, are not
        diminished."
      ],
      "summary": "Enhanced resilience or carrying distance and durability proper-
      ties are needed without diminishing playability characteristics such as feel,
      click, and spin.",
      "evaluation_parameter_1": "resilience or carrying distance",
      "evaluation_parameter_2": "playability characteristics",
      "triz_parameter_1": "'15 - Durability of Moving Object'",
```

```
      "triz_parameter_2": "'33 - Convenience of Use'",
      "action_parameter": "multi-layer covers"
    }
  ]
}
```

## 5 Conclusion and Future Work

This paper showed that GPT-4, as an example of state-of-the-art LLMs, shows promising abilities to extract technical contradictions from patents. Compared to the state-of-the-art of inventive concepts extraction, namely sentence classification with finetuned BERT derivatives and complicated multi-stage approaches based on the same, our approach can easily integrate multiple extraction steps in a single prompt and works "out-of-the-box" with no need for finetuning. The results indicate that a zero-shot *gpt-4-1106-preview* is at least on par with the best performing fine-tuned sentence classifiers for contradiction extraction. Despite of its simplicity, the LLM-approach is not limited to the extraction of sentences, but can also summarize technical contradictions, identify their parameters, and assign them to generic TRIZ engineering parameters.

Weaknesses of the presented approach are a possible lack of reproducibility and the dependency on a commercial black box LLM with comparatively high costs when it comes to the processing of big data like patents.

Future work can focus on the usage of open-source LLM alternatives to GPT-4 and on further result improvement through Prompt Engineering. Furthermore, an extension to more "contradiction-related features", like the assignment of patents to one (or more) of the 40 TRIZ inventive principles is easy to imagine.